\documentclass[runningheads]{llncs}

\usepackage{eccv}

\usepackage[linesnumbered,ruled,vlined]{algorithm2e}
\usepackage{eccvabbrv}

\usepackage{graphicx}
\usepackage{booktabs}
\usepackage{threeparttable}
\usepackage[misc]{ifsym}
 
\usepackage{colortbl} % 导入 colortbl 宏包
\usepackage[accsupp]{axessibility}  % Improves PDF readability for those with disabilities.

\usepackage{amssymb}
\usepackage{amsmath,bm}
\usepackage{enumitem, nicematrix}
\usepackage{multirow}
\newcommand{\STAB}[1]{\begin{tabular}{@{}c@{}}#1\end{tabular}}

\usepackage[listings]{tcolorbox}
\usepackage{wrapfig}

\newtheorem{assumption}[theorem]{Assumption}
\definecolor{blue3}{HTML}{3b75af}

\definecolor{Gray}{gray}{0.94}

\newcommand{\na}{\textsf{n/a}}

\newcommand{\hB}{\mathcal{B}}

\newcommand{\hD}{\mathcal{D}}

\newcommand{\hK}{\mathcal{K}}
\newcommand{\hL}{\mathcal{L}}

\newcommand{\hX}{\mathcal{X}}
\newcommand{\hY}{\mathcal{Y}}

\newcommand{\bI}{\mathbb{I}}
\newcommand{\bE}{\mathbb{E}}
\newcommand{\bR}{\mathbb{R}}

\newcommand{\vx}{{\bf x}}

\newcommand{\vzero}{{\bf 0}}

\newcommand{\vU}{{\bf U}}

\newcommand{\vtheta}{{\boldsymbol \theta}}

\newcommand{\vxi}{{\boldsymbol \xi}}

\newcommand{\valpha}{{\boldsymbol \alpha}}
\usepackage{hyperref}

\usepackage{orcidlink}

\definecolor{fhcolor}{rgb}{0.523, 0.235, 0.625}

\newcommand{\change}[1]{{#1}}
\begin{document}

\title{\change{Learning Scalable} Model Soup on a Single GPU: An Efficient Subspace Training Strategy}

\titlerunning{Memory Efficient Hyperplane Learned Soup (MEHL-Soup)}

\author{Tao Li\inst{1,*},
Weisen Jiang\inst{2,4,*},
Fanghui Liu\inst{3}, 
Xiaolin Huang$^{(\textrm{\Letter})}$\inst{1},
James T. Kwok\inst{2}
}

\authorrunning{T.~Li et al.}

\institute{Department of Automation, Shanghai Jiao Tong University, China
\and
Department of Computer Science and Engineering, The Hong
Kong University of Science and Technology,  Hong
Kong
\and
Department of Computer Science, University of Warwick, United
Kingdom
\and Department of Computer Science and Engineering, Southern University of Science and Technology, China \\
% \url{http://www.springer.com/gp/computer-science/lncs} 
% \and
\email{\{li.tao,xiaolinhuang\}@sjtu.edu.cn},
\email{waysonkong@gmail.com}
\\
\email{fanghui.liu@warwick.ac.uk},
\email{jamesk@cse.ust.hk}
}

\renewcommand{\thefootnote}{*}
\footnotetext[1]{Equal contribution.
Work done when Tao was a visiting student at HKUST.
}

\maketitle

\setcounter{footnote}{0} 
\renewcommand{\thefootnote}{\arabic{footnote}}

\begin{abstract}
Pre-training followed by fine-tuning is widely adopted among practitioners. The performance can be improved by ``model soups''~\cite{wortsman2022model} via exploring various hyperparameter configurations.
The Learned-Soup, a variant of model soups, significantly improves the performance but suffers from substantial memory and time costs due to the requirements of (i) having to load all fine-tuned models simultaneously, and (ii) a large computational graph encompassing all fine-tuned
models.
In this paper, 
we propose \textbf{M}emory \textbf{E}fficient \textbf{H}yperplane \textbf{L}earned Soup (MEHL-Soup) to tackle this issue by formulating the learned soup as a hyperplane optimization problem
and introducing block coordinate gradient descent to learn the mixing coefficients.
At each iteration,
MEHL-Soup only needs to load a few fine-tuned models and build a computational graph with one combined model. 
We further extend {MEHL-Soup} to {MEHL-Soup+} in a layer-wise manner. 
Experimental results on various ViT models and data sets show that
MEHL-Soup(+) outperforms Learned-Soup(+) in terms of test accuracy, and also reduces memory usage by more than $13\times$. 
Moreover,
MEHL-Soup(+) can be run on a single GPU and achieves $9\times$ speed up in
soup construction compared with the Learned-Soup. 
\change{The code is released at \url{https://github.com/nblt/MEHL-Soup}.
}

\end{abstract}

\section{Introduction}
\label{sec:intro}

Pre-training followed by fine-tuning is a widely adopted training pipeline for deep neural networks~\cite{girshick2014rich,yosinski2014transferable,kornblith2019better,kolesnikov2020big, yu2024metamath}. Typically, one starts with a large model pre-trained on an extensive collection of datasets and then fine-tunes multiple models with various hyperparameter configurations to seek better performance. To maximize the benefits of these fine-tuned models, the concept of ``\emph{model soups}''~\cite{wortsman2022model} has been introduced to selectively average the weights of these models for an improved souping model
while keeping the inference efficiency as a single model. 
Model soups have achieved significant success and widely used in various domains, such as out-of-distribution performance~\cite{wortsman2022model,rame2022diverse,chronopoulou2023adaptersoup}, reinforcement learning~\cite{rame2024warm}, model pruning~\cite{zimmer2024sparse,yin2023lottery}, and adversarial robustness~\cite{croce2023seasoning,huang2023adversarial}.
 
There are two representative categories of model soups:
(i) \emph{greedy soup}~\cite{wortsman2022model,rame2022diverse,chronopoulou2023adaptersoup}, in which fine-tuned models are added to the soup sequentially in a greedy order; and
(ii) \emph{learned soup}~\cite{wortsman2022model,li2022trainable}, in which models are mixed by coefficients learned from a validation set.
Greedy soup is simple and effective, but may not explore the full potential of all fine-tuned models
as the models in the soup are equally averaged while others are discarded~\cite{suzuki2022model}.
The learned soup, which learns the soup's mixing coefficients via gradient-based optimization on the validation set, is more general and achieves better performance (e.g., test accuracy) in practice~\cite{wortsman2022model,huang2023adversarial}. 
However, the learned soup
suffers from a heavy burden in computation and memory since it requires loading \textit{all} models into memory simultaneously and building a computational graph on \textit{all} models.
For example, Learned-Soup requires more than 200GB of memory for averaging 72 fine-tuned ViT-B/32 models~\cite{radford2021learning}, and thus the training process has to be conducted in CPU memory~\cite{wortsman2022model}, which is time-consuming.
Hence, Learned-Soup is inefficient in both memory and computation, hindering its application to large models.

In this paper, we develop a scalable and efficient approach to learning the model soup. It works well under limited computational resources, even on a single GPU.
We formulate the learned soup as a subspace learning problem and propose a hyperplane optimization objective, which only requires a computational graph on the combined model.
Furthermore, we introduce block coordinate gradient descent~\cite{tseng2009block,nesterov2012efficiency,wright2015coordinate}
to optimize the mixing coefficients,
where only a mini-batch of models needs to be loaded into memory at each iteration.
This 
not only scales well but also achieves better performance as the introduced stochasticity benefits generalization~\cite{camuto2021fractal,lei2023stability,smith2020generalization}. 

The proposed \textbf{M}emory-\textbf{E}fficient \textbf{H}yperplane \textbf{L}earned Soup (MEHL-Soup) maintains memory and time efficiency while benefiting significant performance improvement from trainable coefficients.
Furthermore,
it is extended in a layer-wise manner (MEHL-Soup+) for boosting performance.
To be specific, our main contributions can be summarized as follows:

\change{
\begin{itemize}
    \item We propose MEHL-Soup(+), a computation- and memory-efficient approach to learning the mixing coefficients of model soup based on a novel hyperplane optimization objective, which allows for learning extrapolated coefficients.
    \item We adopt block coordinate gradient descent to enable training of the model soup scalable and memory-efficient, which can be run on a single GPU. Convergence of MEHL-Soup(+) is also established. 
    % \item We propose MEHL-Soup(+), a computation- and memory-efficient approach to learning the mixing coefficients of model soup, which is scalable and can be run on a single GPU.
    % \item Our MEHL-Soup(+) only needs to load a mini-batch of models for each iteration and build a computational graph on the combined model. It is much more efficient than Learned-Soup, which loads all fine-tuned models and builds a large computational graph on all of them. Convergence of MEHL-Soup(+) is also established. 
    \item Experimental results show that MEHL-Soup(+) brings $13\times$ 
    reduction in memory and 
    $9\times$ in soup construction time compared with Learned-Soup(+) along with consistently better performance. 
    \change{
    Moreover, our findings reveal that compared to Greedy-Soup, MEHL-Soup(+) substantially reduces the cost of fine-tuning and exhibits lower sensitivity to top-performing fine-tuned models, making it more preferable in practice.}
\end{itemize}
}

\section{Related Work}

\noindent
\textbf{Weight Averaging} is a widely used technique in deep learning and optimization for improving generalization~\cite{izmailov2018averaging,szegedy2016rethinking,wortsman2022model,rame2022diverse} and convergence~\cite{hunter1986exponentially,li2022trainable,melis2022two,kaddour2022stop,sanyal2023early}.
Along the same training trajectory, Izmailov~et al.~\cite{izmailov2018averaging} show that averaging the weights at the latter stage of training accompanied with a constant learning rate schedule can significantly improve generalization.
Kaddour~et al.~\cite{kaddour2022stop} introduce the latest weight averaging (LAWA) to accelerate the convergence of training. 
% Li et al.~\cite{li2022trainable} introduce trainable weight averaging (TWA) to further improve the efficiency but face severe memory burden. 
In the context of combining weights from different training trajectories,
Wortsman~et al.~\cite{wortsman2022model} propose to selectively combine multiple fine-tuned models in the parameter space to boost performance.
The combined model is also called a \textit{model soup} \cite{wortsman2022model}.
In inference, only the model soup needs to be deployed and served. Thus, it is computation- and memory-efficient compared with serving all fine-tuned models to combine models in the output space (i.e., model ensemble~\cite{zhou2012ensemble}).
Model averaging has achieved promising performance in a wide variety of applications, including federated learning~\cite{chen2023fedsoup}, robust training~\cite{cai2023robust, rebuffi2022revisiting}, and multi-task training~\cite{ilharco2023editing, ortizjimenez2023task, liu2023tangent}, open-vocabulary recognition~\cite{ilharco2022patching}, and language models alignment~\cite{rame2024warm}.

\noindent
\textbf{Subspace Training.}
Recent studies~\cite{li2018measuring, gressmann2020improving, li2022low, Wortsman2021,li2022subspace} show that
neural networks can be learned in a tiny subspace.
The subspace can be constructed from random basis~\cite{gressmann2020improving, li2018measuring}, training dynamics~\cite{li2022low}, models fine-tuned from a pre-trained model~\cite{wortsman2022model}, and multiple task-specific models~\cite{jiang2022subspace}.
Existing subspace training methods require loading all models into the memory and constructing a computational graph on all of them, leading to the scalability issue for large models.
In contrast, the training strategy proposed in this work only needs a computational graph on the combined model and loads a mini-batch of models. 

\noindent
\textbf{Preliminary.}
Let $\hX \subseteq \mathbb{R}^d$ be a compact metric space and $\hY \subseteq \mathbb{N}$ be the label space for classification. 
The training data $\hD^{\text{tr}} = \{(\vx_i, y_i) \}_{i=1}^n$
and validation data $\hD^{\text{vl}}$
are drawn from an unknown probability distribution on $\hX \times \hY$.
We aim at seeking a hypothesis (i.e., deep network in this work) $f: \hX \to \hY$ such that $f(\vx,\vtheta)$, parameterized by $\vtheta$,
is a good approximation of the label $y$ corresponding to a new sample $\vx \in \hX$.
The loss function $\ell(f(\vx,\vtheta),y)$ (e.g., cross-entropy loss)
measures the discrepancy between the prediction $f(\vx,\bm \theta)$ and label $y$.
The generalization performance can be evaluated by the expected risk $\mathbb{E}_{(\bm x, y)} \ell(f(\vx,\vtheta),y)$.

Let $\vtheta=\texttt{fine-tune}(\hD^{tr}, \vtheta_0, h)$ be the model parameters obtained through fine-tuning on the training data $\hD^{tr}$ with pre-trained weights $\vtheta_0$ and a specific hyperparameter configuration $h$. 
This hyperparameter configuration can encompass aspects such as the learning rate, weight decay, data augmentation, and random seed, among others~\cite{wortsman2022model}. 
For a set of $K$ hyperparameter configurations $\{ h_k \}_{k=1}^K$, let $\vtheta_k=\texttt{fine-tune}(\hD^{tr}, \vtheta_0, h_k)$ denote the model parameters obtained through fine-tuning with the $k$th configuration $h_k$.
Accordingly, the fine-tuned models $\{\vtheta_k\}_{k=1}^K$
can be combined together to enhance generalization performance.
We review representative model soup methods introduced by Wortsman et al.~\cite{wortsman2022model}

\textbf{Uniform-Soup ($\vtheta_{\text{US}}$)} (or SWA~\cite{izmailov2018averaging}) is the most straightforward method to obtain a model soup by uniformly averaging all fine-tuned models:
\begin{align}
    \vtheta_{\text{US}} = \frac{1}{K}\sum_{k=1}^K \vtheta_k.
\end{align}
This simple averaging approach may not always enhance generalization performance, as the hyperparameters used for fine-tuning are typically randomly searched and can exhibit significant diversity.

\textbf{Greedy-Soup} improves Uniform-Soup by selectively averaging a subset of models. 
Specifically, it first sorts the fine-tuned models according to their validation accuracies and then sequentially adds models to the soup if the validation performance of the soup is improved. 
Greedy-Soup empirically outperforms uniform soup and is adopted by practitioners.

\textbf{Learned-Soup ($\vtheta_{\text{LS}}$)} constructs a model soup by learning coefficients to combine the fine-tuned models. 
The objective is formulated as follows:
\begin{align}
    \begin{split}
    \min_{\valpha} \quad & \hL(\vtheta_{\rm LS};\hD^{\text{vl}})\\
\rm{s.t.}~~~ &\vtheta_{\rm LS}=\alpha_1 \vtheta_1 + \alpha_2 \vtheta_2 + \cdots + \alpha_K \vtheta_K,\\
&\alpha_1+\alpha_2+\cdots+\alpha_K=1, \\
&\alpha_k \in [0, 1], \;\; k=1,\dots,K.
    \label{equ:goal}
\end{split}
\end{align}
Solving a constrained optimization problem is challenging.
In practice, Wortsman et al. \cite{wortsman2022model} 
resolve this problem by introducing a parameterization of $\valpha$ using the softmax function, ensuring that each $\alpha_k$ lies in $[0,1]$ and their values sum up to one.
The Learned-Soup can be further enhanced by considering the layer-wise structure of deep networks and assigning individual coefficients to each layer. 
The Learned-Soup is general but
needs to load \textit{all} fine-tuned models in the memory and build the computational graph on \textit{all} fine-tuned models for learning $\valpha$, which is infeasible due to memory and computation considerations.
Hence, the
Learned-Soup
is rarely used in practice compared with the Greedy-Soup. 
In this work,
we propose an efficient algorithm to address the memory and computational issues of the Learned-Soup.
Different from the Learned-Soup,
the proposed algorithm only needs to build the computational graph on the combined model
and load a mini-batch of fine-tuned models.

\section{Methodology}

In this section, we present our
memory-efficient learned soup approach. 
We start by formulating the optimization target as a hyperplane optimization problem
(\cref{sec:subspace}),
then introduce our efficient coefficient optimization method
(\cref{sec:opt}),
and finally employ block coordinate gradient descent to avoid loading all fine-tuned models for learning coefficients
(\cref{sec:cgd}).

\subsection{Subspace Learning: Hyperplane Optimization Target}
\label{sec:subspace}

For learning the mixing coefficients, the Learned-Soup~\cite{wortsman2022model} proposes to learn from a probability simplex using the softmax operation.
Specifically, 
a softmax layer is applied to the
learnable variables to
output the coefficient vector $\valpha$, which naturally
satisfies the convex-hull constraints: $\alpha_k \in [0, 1]$ and $\sum_{k=1}^K \alpha_k=1$.
However, recent studies~\cite{rebuffi2022revisiting,croce2023seasoning} show that interpolation within the convex hull may lead to sub-optimal performance, and extrapolation is more general and can perform better. 

Previous computations of the extrapolation weights are normally based on grid search for just two models~\cite{rebuffi2022revisiting,croce2023seasoning}.
It becomes less efficient when merging numerous fine-tuned models as the solution space grows exponentially with the number of models involved.
In this work, we develop a novel approach based on subspace learning to seek mixing coefficients that incorporate extrapolation. 
This relaxes the solution space from a convex hull to a hyperplane spanned by the fine-tuned models $\vtheta_k$'s.
Formally,
the objective in \cref{equ:goal} is changed to:
\begin{align}
\begin{split}
    \min_{\valpha} \quad &\hL(\vtheta_{\star}; \hD^{\text{vl}})\\
    \rm{s.t.}~~~ &\vtheta_{\star}=\bar{\vtheta}+\alpha_1 (\vtheta_1-\bar{\vtheta}) + \alpha_2 (\vtheta_2-\bar{\vtheta}) + \cdots + \alpha_K (\vtheta_K-\bar{\vtheta})\,,
    \label{equ:ourtarget}
\end{split}
\end{align}
where $\bar{\vtheta}:=\frac{1}{K}\sum_{k=1}^K{\vtheta}_k$. 
One can see that the constraints on 
$\valpha$ have been removed to allow for extrapolation, and the combined model $\vtheta_{\star}$ lies in a hyperplane spanned by $\{ \vtheta_k\}_{k=1}^K$. 
\change{Note also that
    Eq. (\ref{equ:ourtarget}) uses $\vtheta_i-\bar{\vtheta}$  instead of $\vtheta_i$,
which reduces the correlations among $\vtheta_i$'s,
which also enables better performance (please refer to Appendix~D for more details).} 
We refer to this approach as the Hyperplane Learned Soup (HL-Soup) 
to distinguish it from previous methods focusing on the convex hull. 

Note that 
the mixing coefficients still sum to one, allowing for numerical stability.
To see that, $\vtheta_{\star}$ in \cref{equ:ourtarget} can be equivalently rewritten as:
\begin{align}
    \vtheta_{\star}=\sum_{k=1}^K \left(\frac{1}{K} + \alpha_k-\frac{1}{K}\sum_{k'=1}^K\alpha_{k'}\right) \vtheta_k\,,
\end{align}
which leads to the identity
$
    \sum_{k=1}^K \left(\frac{1}{K} + \alpha_k-\frac{1}{K}\sum_{k'=1}^K\alpha_{k'}\right) =1
$.
By doing so, we can effectively incorporate extrapolation into the weight combination, leading to improved performance.

To further enhance the representation ability of HL-Soup, we introduce a layer-wise mixing scheme called HL-Soup+ as follows: 
\begin{align}
\begin{split}
    \min_{\valpha} \quad & \hL(\vtheta_{\star}; \hD^{\text{vl}}) \\
    \rm{s.t.}~~~ &\vtheta^{(l)}_{\star}=\bar{\vtheta}^{(l)} \!+\!\alpha_1^{(l)}(\vtheta_1^{(l)}\!-\!\bar{\vtheta}^{(l)}) \!+\! \alpha_2^{(l)}(\vtheta_2^{(l)}\!-\!\bar{\vtheta}^{(l)})\! +\! \cdots \!+\! \alpha_K^{(l)} (\vtheta_K^{(l)} \!-\!\bar{\vtheta}^{(l)})\,,\\
    &l\in\{0,1,\dots,L\}\,. \label{equ:ourtarget-layer} 
\end{split}
\end{align}
The use of layer-wise averaging facilitates a more precise manner of model averaging, thereby enhancing the utilization of fine-tuned models and resulting in better performance.
In \cref{sec:further_comparison}, we will show that in order to achieve similar test performance, the layer-wise approach requires fewer fine-tuned models compared to the greedy soup, which leads to significant computational savings in the fine-tuning stage.

\subsection{Efficient Coefficient Optimization}
\label{sec:opt}
How to efficiently optimize the mixing coefficients
$\valpha$ in \cref{equ:ourtarget-layer} is one of the main targets of this work.
The classical approach is to construct a computational graph by wrapping all the fine-tuned models and then compute the respective gradient under forward and backward propagation~\cite{wortsman2022model}.
This method is memory-inefficient
due to (i) multiple memory footprints for the internal computational state to build the computational graph; and (ii) all fine-tuned models need to be loaded.

Indeed, the memory issue associated with optimizing the mixing coefficients has been acknowledged as an ongoing challenge~\cite{wortsman2022model}.
To address this, Wortsman et al.~\cite{wortsman2022model} initially propose combining models in CPU rather than GPU,
as CPU generally offers larger memory capacities. However, training on the CPU can be substantially slower compared to training on the GPU. Additionally, despite this adjustment, the memory issue is not entirely resolved since the memory capacity of the CPU still remains limited.

In this work, by using the proposed hyperplane optimization, we can conveniently derive the gradient with respect to $\alpha_k$
as follows:
\begin{align}
% \begin{split}
    \nabla_{\alpha_k} \hL(\vtheta_{\star};\hD^{\text{vl}})
    = \nabla_{\vtheta_{\star}}^\top \hL(\vtheta_{\star};\hD^{\text{vl}}) \nabla_{\alpha_k} \vtheta_{\star}  
    = \nabla_{\vtheta_{\star}}^\top \hL(\vtheta_{\star};\hD^{\text{vl}})
    (\vtheta_k - \bar{\vtheta})\,.
    \label{equ:grad-proj}
% \end{split}
\end{align}
To simplify notations, we consider the non-layer-wise case here.  Extension to the layer-wise case is straightforward.
The first component $\nabla_{\vtheta_{\star}}^\top \hL(\vtheta_{\star};\hD^{\text{vl}})$ in
    Eq.~(\ref{equ:grad-proj})
is shared across all fine-tuned models, and
thus only one computational graph on the model soup $\vtheta_{\star}$ is required.
The second component, $\vtheta_k - \bar{\vtheta}$, is specific to the $k$th fine-tuned model. 
Hence, 
$\{\nabla_{\valpha_k} \hL(\vtheta_{\star};\hD^{\text{vl}})\}_{k=1}^K$ only needs the computational graph on $\vtheta_{\star}$,
which is affordable for a single GPU.
Thus, this 
addresses the additional memory burden
caused by computational graph construction, which typically requires memory that is multiple times the model size (\cref{sec:main-exp}).
Moreover,
the simple inner product operation in \cref{equ:grad-proj} is particularly advantageous for leveraging GPU acceleration.
The remaining memory burden is caused by caching all fine-tuned models and will be resolved in the next section.

\subsection{Block Coordinate Gradient Descent}
\label{sec:cgd}

To avoid caching all the fine-tuned models, we borrow the classical idea of block coordinate gradient descent (BCGD)~\cite{tseng2009block,nesterov2012efficiency,richtarik2016parallel,richtarik2016distributed} for stochastic approximation: We randomly select and update a block of variables by gradient descent at each iteration while keeping the remaining variables fixed. 
This allows for learning the coefficients without caching all fine-tuned models in memory.

Formally, at iteration $t$,
we sample a mini-batch of $b$ coordinates $\hK_t\!=\!\{t_1,\dots, t_b\}$ \\ $\subseteq \{1,\dots,K\}$ 
and update
$\{\alpha_{k,t}: k\in \hK_t\}$ while keeping $\{\alpha_{k,t}: k\notin \hK_t\}$ unchanged.
Obviously, computation for 
$\{\nabla_{\alpha_{k,t}} \hL(\vtheta_{\star};\hD^{\text{vl}}): k \notin \hK_t\}$ at iteration $t$ is not included in BCGD, and
only the fine-tuned models corresponding to the chosen coordinates in $\hK_t$ are considered. 
To be specific, let $\vtheta_{{\star}, t}$ be the model soup at iteration $t$.
By \cref{equ:grad-proj}, the update rule for one gradient descent step of the mixing coefficients can be written as:
\begin{align}
% \begin{split}
    \alpha_{k, t+1}  = 
     \begin{cases}
     \alpha_{k, t}  - \eta \nabla_{\vtheta_{{\star}, t} }^\top \hL(\vtheta_{{\star}, t};\hD^{\text{vl}})
    (\vtheta_k - \bar{\vtheta} )\, &\text{if } k \in \hK_t \\
    \alpha_{k, t} &\text{if } k \notin \hK_t
     \end{cases}
    \label{eq:temp-alpha-a1}
    % &= \,, \\
    % \alpha_{k, t+1}  &= \alpha_{k, t} \,, \quad\quad\quad\quad\quad\quad\quad\quad\quad\quad\quad\quad~~
    % \text{for } k \notin \hK_t\,. 
    % \end{split}
\end{align}
Using the updated coefficients,
we update the model soup as:
\begin{align}
\begin{split}
    \vtheta_{{\star}, t+1}&=\bar{\vtheta} +\sum_{k=1}^K\alpha_{k, t+1}(\vtheta_{k}\!-\!\bar{\vtheta}) \\
    % &=\bar{\vtheta} +\sum_{k\notin\hK_e}\alpha_{k, e}^{(l)}(\vtheta_{k}^{(l)}\!-\!\bar{\vtheta}^{(l)}) +\sum_{k\in\hK_e}\alpha_{k, e+1}^{(l)}(\vtheta_{k}^{(l)}\!-\!\bar{\vtheta}^{(l)}) \\
    &= \vtheta_{{\star}, t} - \underbrace{\sum_{k\in\hK_t}\alpha_{k, t}(\vtheta_{k}\!-\!\bar{\vtheta}) +\sum_{k\in\hK_t}\alpha_{k, t+1}(\vtheta_{k}\!-\!\bar{\vtheta})}_{\text{independent of $\{\vtheta_{k'}: k'\notin \hK_t\}$}}\,.\end{split}
\end{align}
Therefore, the new soup $\vtheta_{{\star}, t+1}$ can be constructed from the previous soup $\vtheta_{{\star}, t}$ and a weighted combination of the chosen fine-tuned models. 
In total, we only need to cache $b+1$ models (one model soup and $b$ fine-tuned models), which is much more memory-efficient than the Learned-Soup\cite{wortsman2022model}
that requires caching $K+1$ models. 
Together with \cref{eq:temp-alpha-a1},
the mixing coefficients can be learned
without the necessity of loading all models,
which effectively resolves the memory issue from ${O} (KD)$ to ${O} (bD)$, where $D$ is the number of model parameters.

The process of loading/unloading a mini-batch of models into memory at each iteration is time-consuming.
To improve efficiency,
at each (outer) iteration $t$,
we load the chosen models $\{\vtheta_k: k \in \hK_t\}$ and update the corresponding coefficients $\{\valpha_{k}: k \in \hK_t\}$ for $J$ successive (inner) iterations.
The whole procedure of learning layer-wise mixing coefficients,
called a Memory-Efficient training algorithm for a Hyperplane Learned Soup (denoted MEHL-Soup+), is shown in
Algorithm \ref{alg:mhls}.
MEHL-Soup+ is memory-efficient and scalable to large and numerous models.

\begin{algorithm}[!t]
\caption{MEHL-Soup+.} \label{alg:mhls}
\label{algorithm}
\KwIn{potential soup ingredients $\{ {\bm \theta}_1,\cdots,{\bm \theta}_K\}$, outer iterations $T$, 
inter iterations $J$, model mini-batch size $b$, \#layers $L$, learning rate $\eta$}
\KwOut{learned soup ${\bm \theta}_{\star}$}  
\BlankLine
$\bar{\vtheta}=\frac{1}{K}\sum_{k=1}^K\vtheta_k$; \\ initialize $\vtheta_{\star, 0} = \bar{\vtheta}$, 
 $\valpha_{k,0}=\vzero$ for $k=1,\dots, K$; \\
 \For{$t=1, \dots, T$}{
 sample a mini-batch of $b$ coordinates $\hK_t=\{t_1, \dots, t_b\}$; \\
 load the fine-tuned models $\{\vtheta_{k}\}_{k\in \hK_t}$; \\
 $\vtheta_{\text{fix}}^{(l)} \!=\! \textsf{stop\_gradient}\!\left(\!\vtheta_{\star, (t-1)J}^{(l)} \!-\!\! \sum_{k\in \hK_t} \!\!\alpha_{k, (t-1)J}^{(l)} (\vtheta_k^{(l)} \!\!-\! \bar{\vtheta}^{(l)})\!\right)\!$ for $l\!=\!0,\dots, L$;\!\!\!\!\!\!\!\!\!  \\
 \For{$j=1,\dots,J$}{
 compute \#iterations $i=(t-1)J+j-1$; \\
 ${\vtheta}_{\star, i}^{(l)}=\vtheta_{\text{fix}}^{(l)}+\sum_{k\in \hK_t} \alpha_{k, i}^{(l)} (\vtheta_{k}^{(l)} - \bar{\vtheta}^{(l)})$ for $l=0,\dots, L$;\\
 sample a mini-batch validation data $\hB_{i}$ from $\hD^{\text{vl}}$;\\
 calculate gradients $  \{\nabla_{\valpha_{k, i}} \hL(\vtheta_{\star,i};\hB_{i}):k \in \hK_t \}$ by \cref{equ:grad-proj} for $l\!=\!0, \dots, L$;\!\!\!\!  \\
 \For{$k=1,\dots, K$}{
 \If{$k\in \hK_t$}{
 $\valpha_{k,i+1}=\valpha_{k, i} -\eta \nabla_{\valpha_{k, i}} \hL(\vtheta_{\star,i};\hB_{i})$;}
\Else{
 $\valpha_{k,i+1} =\valpha_{k, i} $;}
 }
 }
  ${\vtheta}_{\star, tJ}^{(l)}=\vtheta_{\text{fix}}^{(l)}+\sum_{k\in \hK_t} \alpha_{k, tJ}^{(l)} (\vtheta_{k}^{(l)} - \bar{\vtheta}^{(l)})$ for $l=0,\dots, L$;
 }
\Return $\vtheta_{\star, TJ}$.
\end{algorithm}

\subsection{Convergence Analysis}
In this section,
we study the convergence of Algorithm~\ref{alg:mhls}.
We first make some assumptions that are standard in stochastic optimization~\cite{ghadimi2013stochastic, Reddi2016,jiang2023an,li2024friendly,lirevisiting,si2024practical}.

\begin{assumption}[Smoothness]
    $\hL(\valpha; \hD^{\text{vl}})$ is $\beta$-smooth in $\valpha$, i.e., 
    $$\| \nabla_{\valpha} \hL(\valpha; \hD^{\text{vl}}) - \nabla_{\valpha'} \hL(\valpha'; \hD^{\text{vl}}) \| \leq \beta \| \valpha - \valpha'
     \|.$$
\end{assumption}
\begin{assumption}[Bounded variance] 
There exists $\sigma>0$ such that
    $$\bE_{(\vx, y) \sim \hD^{\text{vl}}} \left\| \nabla_{\valpha} \ell(f(\vx; \valpha), y) - \nabla_{\valpha} \hL(\valpha; \hD^{\text{vl}}) \right\|^2 \leq \sigma^2.$$
\end{assumption}
\begin{theorem} \label{thm:convergence}
    If the learning rate 
    $\eta \leq \min \{\frac{1}{\beta}, \frac{1}{\sqrt{T}}\}$, 
    Algorithm~\ref{alg:mhls} satisfies 
    \begin{align}
         \min_{1\leq t \leq T} \bE  \|\nabla_{\valpha_{\cdot, tJ}}\hL(\valpha_{\cdot, tJ}; \hD^{\text{vl}}) \|^2    \!\leq \! \frac{2K\left(\bE \hL(\valpha_{\cdot, 1}; \hD^{\text{vl}}) \!-\! \bE \hL(\valpha_{\cdot, TJ}; \hD^{\text{vl}})\right)}{b\sqrt{T}} \!+\!  \frac{\beta J  \sigma^2 K}{b\sqrt{T}}\,, \nonumber
    \end{align}
    where the expectation is taken over the
    random mini-batch of samples
    and models.
\end{theorem}
The proof can be found in Appendix~A.
The $O\left(\frac{1}{\sqrt{T}}\right)$ speed matches the convergence rate in~\cite{ghadimi2013stochastic}.
Moreover,
we can see that increasing the batch size $b$ of models decreases the upper bound.
As a large $b$ intensifies the burden on memory,
there is a trade-off between convergence rate and memory-efficiency.
When $b=K$,
it reduces to the learned soup with extrapolated mixing weights.

\section{Experiments} 

In this section, 
experiments are performed to demonstrate the efficiency and effectiveness of the proposed methods. We begin by evaluating various model soup methods on the ImageNet dataset. We then conduct experiments specifically targeting a larger model.
Finally, we provide a detailed comparison between greedy and learned soup methods and ablation studies. 
\change{
More experiments on ResNet~\cite{he2016deep} can be found in Appendix C.
}

\subsection{Experiments on ViT-B/32}
\label{sec:main-exp}

{\bf Setup.} 
Following~\cite{wortsman2022model},
we perform experiments on the ImageNet~\cite{Russakovsky2015} using the pre-trained CLIP ViT-B/32.
We use the publicly available fine-tuned models provided by~\cite{wortsman2022model}. 
They are obtained by a random hyperparameter search over the learning rate, weight decay, training epochs, label smoothing, and data augmentation, resulting in a total of 72 fine-tuned models.  
Training details can be found in Appendix~B.

\noindent
\textbf{Baselines.}
The proposed HL-Soup(+) and MEHL-Soup(+) are compared with (i)
Best individual model with the highest accuracy on the validation set, 
(ii)
Uniform-Soup~\cite{izmailov2018averaging}, which averages all model parameters uniformly, 
(iii) Greedy-Soup~\cite{wortsman2022model}, which greedily adds models to the soup to improve validation accuracy, 
(iv)
Learned-Soup~\cite{wortsman2022model}, which learns coefficients to combine models,
and (v) Ensemble, which combines
the outputs of all fine-tuned models by aggregating their logit outputs.
All methods use the same fine-tuned models and validation set.
We utilize the official code provided by~\cite{wortsman2022model} for reproducing Greedy-Soup and Learned-Soup. 

\begin{table}[!t]
    \centering
	\setlength{\tabcolsep}{5pt}
    \caption{Comparison of different methods on ImageNet with pre-trained CLIP ViT-B/32. The number of fine-tuned models is 72. We measure the time and memory on a server with one NVIDIA GeForce RTX 4090 GPU and 256 GB RAM. }
    \label{tab:results-imagenet}
    % \vskip -.15in
    \begin{threeparttable}
    {
    \footnotesize 
    \resizebox{.9\textwidth}{!}{
    \begin{tabular}{@{}lccccc@{}}
    \toprule
    &Testing &Time & &Soup &Peak \\
    Method &accuracy  &per &\#epoch &construction &memory   \\
    &(\%) &epoch & &time &burden \\
    \midrule
    Best individual model  &80.38 &- &- &- &- \\
    % &1365s &10 &13650.1s\textsuperscript{a} &9.4GB\textsuperscript{a}\\
    \midrule
    Uniform-Soup~\cite{izmailov2018averaging} &79.97 &- &- &- &-\\
         Greedy-Soup~\cite{wortsman2022model} &81.03 &24s 
         % &24.5s 
         &143\textsuperscript{a} &3501s &3GB\\
        \arrayrulecolor{lightgray}\midrule
        Learned-Soup~\cite{wortsman2022model} &80.88 &1503s &5 &7514s &249GB\textsuperscript{b}\\  
        HL-Soup~(\textbf{ours})  &81.14 &496s &5 &2479s &43GB\textsuperscript{b}\\  
        MEHL-Soup~(\textbf{ours})  &\textbf{81.20} &39s &20\textsuperscript{c} &776s &18GB\\
        \arrayrulecolor{lightgray}\midrule
         Learned-Soup+~\cite{wortsman2022model} &81.39 &1540s &5 &7701s &253GB\textsuperscript{b}\\  
         HL-Soup+~(\textbf{ours})   &81.45 &581s &5 &2903s &44GB\textsuperscript{b}\\  
         MEHL-Soup+~(\textbf{ours})   &\textbf{81.62} &40s &20\textsuperscript{c} &808s{\scriptsize \color{teal} ($\downarrow 9.5\times$)} &19GB{\scriptsize \color{teal} ($\downarrow 13\times$)}\\\arrayrulecolor{lightgray}\midrule
         Ensemble\textsuperscript{d} 
         %(more cost) 
         &81.19 &- &- &- &-\\
        \arrayrulecolor{black}
    \bottomrule
    \end{tabular}}
    }
    \begin{tablenotes}
    % \vspace{0.5mm}
    \scriptsize
      % \item[\textsuperscript{a}] We report the time and memory cost for fine-tuning each individual model.
      \item[\textsuperscript{a}] Greedy-Soup involves two evaluation stages: the first stage evaluates all models and sorts them, while the second stage sequentially adds each model to the soup.
      \item[\textsuperscript{b}] We place the fine-tuned models in CPU memory following~\cite{wortsman2022model} and report CPU memory usage as it is too large to fit into 24 GB GPU memory. 
      \item[\textsuperscript{c}] We use a mini-batch of 18 models and hence the corresponding number of training epochs for MEHL-Soup(+) is  4x longer than those of learned soup and HL-Soup(+).
      \item[\textsuperscript{d}] Ensemble directly utilizes the outputs of all models and does not require soup construction.
      % additional notes
    \end{tablenotes}
    \end{threeparttable}
    % \vspace{-5mm}
    % \vskip -.2in
\end{table}

\noindent {\bf Results.} As can be seen from \cref{tab:results-imagenet},
MEHL-Soup+ achieves the highest accuracy.
Besides, MELH-Soup+ achieves an accuracy gain of 1.24\% over the best individual model, 
demonstrating the effectiveness of learning the model soup. 

\noindent
{\it Comparison with Greedy-Soup.}
The proposed learned soup approaches outperform Greedy-Soup by an accuracy gain of 0.17\% via MEHL-Soup and 0.59\% via MEHL-Soup+, respectively.
Regarding the soup construction time, 
MEHL-Soup(+) is $4\times$ faster than Greedy-Soup.
Recall that
for Greedy-Soup, the number of validation performance evaluations is equal to 
twice the number of fine-tuned models (one on sorting the fine-tuned models and one for performance evaluation after each candidate model is added to the soup).
This can be even larger than the number of training epochs for MEHL-Soup(+), particularly when there are numerous models, e.g., 143 epochs of validation for Learned-Soup(+) and 20 epochs training for MEHL-Soup(+) with 72 models, and thus
MEHL-Soup(+) can be faster.

\noindent {\it Comparison with Learned-Soup.}
MEHL-Soup achieves higher accuracy (+0.32\%) than Learned-Soup.
The source of accuracy improvement
may come from the weight extrapolation (i.e., mixing coefficients outsize (0,1)) is more flexible than the Learned-Soup whose coefficients are constrained in $(0,1)$ due to softmax parameterization.
This observation also agrees with recent findings~\cite{croce2023seasoning, rebuffi2022revisiting} that weight
extrapolation can boost the performance of combining two
models.
For the layer-wise scheme, 
MEHL-Soup+ also performs better than Learned-Soup+.
In particular, note that for the non-layerwise scheme, Learned-Soup is worse than Greedy-Soup while the proposed MEHL-Soup still achieves higher accuracy, confirming that learning mixing coefficients from the hyperplane is better than the probability simplex.

Regarding the soup construction time
and memory burden,
Learned-Soup incurs a significant memory burden, with peak memory reaching as high as 253 GB. 
This is primarily due to the need to build a computational graph on all fine-tuned models, which severely increases the additional memory overhead.
The burden becomes particularly evident when dealing with larger models.
Consequently, the process of combining models has to be carried out in CPU memory, 
significantly slowing down the training speed.
Instead, by our subspace training approach and employing a mini-batch coordinate gradient descent strategy, we successfully reduce the memory burden by $13\times$, allowing efficient training with a single GPU, and then remarkably reduce the corresponding soup construction time by $9.5\times$.

\noindent
{\it Comparison with Ensemble.} 
We observe that our learned approach can significantly outperform the model ensemble (e.g., by +0.43\% with MEHL-Soup+). Note that ensemble typically requires higher inference costs since they involve aggregating the outputs of all models.

In all, the proposed MEHL-Soup(+) addresses the huge memory requirements of previous methods, enabling scalability and efficient execution on a single GPU.
Moreover, MEHL-Soup(+) achieves higher accuracy than Greedy-Soup.

\begin{wrapfigure}{r}{0.46\textwidth}
    \centering
\includegraphics[width=.35\textwidth]{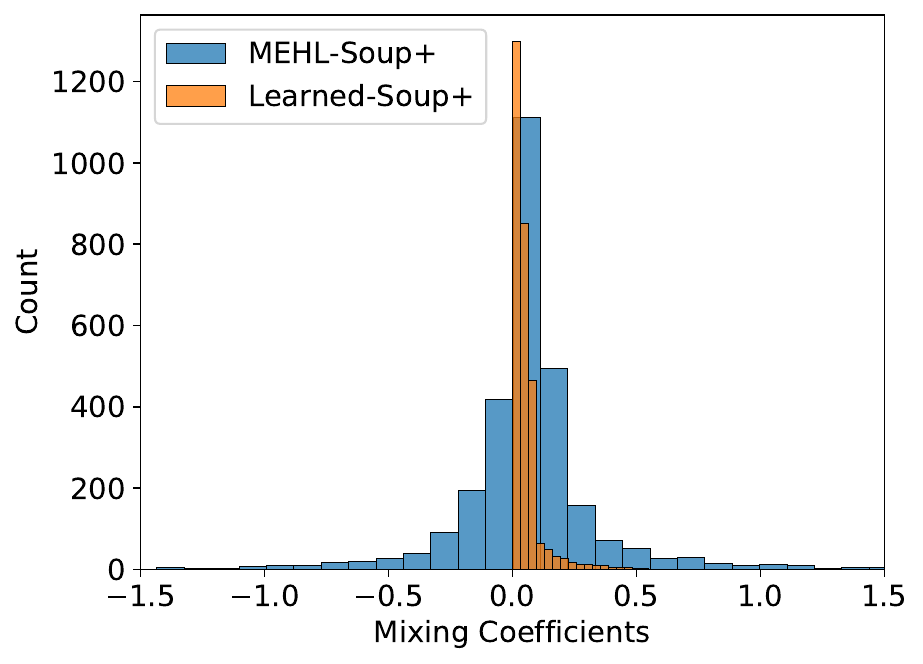}
% \vskip -.15in
    \caption{Distributions of mixing coefficient over all layers and models learned by MEHL-Soup+ and Learned-Soup+
    % . The experiment is conducted 
    on ImageNet with CLIP ViT-B/32.
    }
    \label{fig:coeff-pic}
    % \vskip -.3in
\end{wrapfigure}

\noindent
{\it Visualization of Mixing Coefficients.} 
\cref{fig:coeff-pic}
shows the distributions of mixing  
coefficients for all the layers and fine-tuned models learned by MEHL-Soup+ and Learned-Soup+. As can be seen, MEHL-Soup+ learns extrapolated coefficients (i.e., values outside [0, 1]), but Learned-Soup+ enforces the constraint that coefficients must lie in the range $(0, 1)$. \change{The extrapolated coefficients can lead to better performance as demonstrated in Tables~\ref{tab:results-imagenet} and \ref{tab:results-vitl}.}
Furthermore, as can be seen,
most of the coefficients are close to zero for both methods.

\begin{table}[!t]
    \centering
	\setlength{\tabcolsep}{4pt}
    \caption{Comparison of different methods with pre-trained CLIP ViT-L/14. The number of fine-tuned models is 32 for CIFAR-10/100 and 8 for ImageNet. We measure the time and memory on a server with one NVIDIA GeForce RTX 4090 GPU and 256 GB RAM. ``\na'' means the result is not available due to Out-Of-Memory.}
    \label{tab:results-vitl}
    % \vskip -.15in
    % \vskip -.15in
    % \begin{threeparttable}
    % {
    \footnotesize 
    \resizebox{.9\textwidth}{!}{
    \begin{NiceTabular}{c|lccccc@{}}
    \CodeBefore
	\Body
    \toprule
    &&Testing  &Time & &Soup &Peak \\
    & Method &accuracy &per &\#epoch &construction &memory   \\
    & &(\%) &epoch & &time &burden \\
    \midrule
    \multirow{10}{*}{\STAB{\rotatebox[origin=c]{90}{CIFAR-10}}}  &  
    Best individual model  &98.93 &- &- &- &-\\
    \cmidrule{2-7}
        &Uniform-Soup~\cite{izmailov2018averaging} &99.23 &- &- &- &- \\
         &Greedy-Soup
         ~\cite{wortsman2022model} 
         &99.15 &80s &63 &5046s &6GB\\
        \arrayrulecolor{lightgray}\cmidrule{2-7}
        &Learned-Soup~\cite{wortsman2022model}  & \na & \na & \na & \na &>256GB\\
       & HL-Soup~(\textbf{ours}) &99.25 &1203s &5 &6015s &49GB \\
       & MEHL-Soup~(\textbf{ours}) &99.26 &199s &20 &3976s &23GB\\
        \arrayrulecolor{lightgray}\cmidrule{2-7}
        & Learned-Soup+~\cite{wortsman2022model} & \na & \na & \na & \na &>256GB \\
       &HL-Soup+ (\textbf{ours}) &99.24 &1241s &5 &6205s &49GB\\
        & MEHL-Soup+ (\textbf{ours}) &\textbf{99.27} &199s &20 &3988s &23GB \\
        \arrayrulecolor{lightgray}\cmidrule{2-7}
        &Ensemble 
        &99.14 &- &- &-  &- \\
        \arrayrulecolor{black}
    \midrule
    \multirow{10}{*}{\STAB{\rotatebox[origin=c]{90}{CIFAR-100}}}
    & Best individual model  &92.49 &- &- &- &-\\
    \cmidrule{2-7}
        &Uniform-Soup~\cite{izmailov2018averaging} &93.05 &- &- &- &- \\
        & Greedy-Soup~\cite{wortsman2022model} &93.32 &81s &63 &5084s &6GB\\
        \arrayrulecolor{lightgray}\cmidrule{2-7}
       & Learned-Soup~\cite{wortsman2022model}  & \na & \na & \na & \na &>256GB \\
       & HL-Soup~(\textbf{ours}) &93.41 &1251s &5 &6255s &50GB\\
      &  MEHL-Soup~(\textbf{ours}) &93.52 &200s &20 &4002s &23GB\\
        \arrayrulecolor{lightgray}\cmidrule{2-7}
       &  Learned-Soup+~\cite{wortsman2022model} & \na & \na & \na & \na &>256GB\\
       &HL-Soup+ (\textbf{ours}) &93.59 &1255s &5 &6275s &50GB\\
       &  MEHL-Soup+ (\textbf{ours}) &\textbf{93.70}  &205s &20 &4100s &23GB\\
        \arrayrulecolor{lightgray}\cmidrule{2-7}
        &Ensemble 
        &93.65 &- &- &- &-\\
        \arrayrulecolor{black}
    \midrule
    \multirow{10}{*}{\STAB{\rotatebox[origin=c]{90}{ImageNet}}}
    & Best individual model  &85.48 &- &- &- &-\\
    \cmidrule{2-7}
        &Uniform-Soup~\cite{izmailov2018averaging} &85.11 &- &- &- &- \\
        & Greedy-Soup~\cite{wortsman2022model} &85.64 &473s &15 &7097s &6GB\\
        \arrayrulecolor{lightgray}\cmidrule{2-7}
       & Learned-Soup~\cite{wortsman2022model} &85.20 &7340s &5 &36700s &90GB\\
       & HL-Soup~(\textbf{ours}) &85.70 &950s &5 &4748s &23GB\\
      &  MEHL-Soup~(\textbf{ours})\tabularnote{MEHL-Soup(+) recovers HL-Soup(+) as we use a mini-batch of 8 models, which can fit into a single GPU.} &85.70 &950s &5 &4748s &23GB\\
        \arrayrulecolor{lightgray}\cmidrule{2-7}
       &  Learned-Soup+~\cite{wortsman2022model} &85.53  &7372s &5 &36858s &90GB\\
       &HL-Soup+ (\textbf{ours}) &\textbf{86.03}&1066s &5 &5330s &23GB\\
       &  MEHL-Soup+ (\textbf{ours})\tabularnote{MEHL-Soup(+) recovers HL-Soup(+) as we use a mini-batch of 8 models, which can fit into a single GPU.} &\textbf{86.03}  &1066s &5 &5330s{\scriptsize \color{teal} ($\downarrow 6.9\times$)} &23GB{\scriptsize \color{teal} ($\downarrow 3.9\times$)}\\
        \arrayrulecolor{lightgray}\cmidrule{2-7}
        &Ensemble 
        &86.10 &- &- &- &- \\
        \arrayrulecolor{black}
    \bottomrule
    \end{NiceTabular}
    }
\end{table}

\subsection{Experiments on Larger ViT-L/14} \label{sec:vit-L14}

After resolving the memory issues, the proposed training approach can be used to learn mixing coefficients for combining \textit{larger} models, which is impractical for the previous Learned-Soup method due to the substantial memory requirement. 
To demonstrate this,
we adopt the CLIP ViT-L/14 model\footnote{ViT-L/14 contains 343M parameters, while ViT-B/32 contains only 87M parameters.}
\cite{radford2021learning}
and evaluate on three datasets: CIFAR-10, CIFAR-100, and ImageNet. The fine-tuning and training details can be found in Appendix~B.

\cref{tab:results-vitl}
shows the results.\
As can be seen, MEHL-Soup+ consistently achieves higher accuracy 
than Greedy-Soup (+0.12\%
on CIFAR-10, +0.38\% on CIFAR100, and +0.39\% on ImageNet). Moreover, when there are 32 fine-tuned models, Learned-Soup requires over 256GB of memory and cannot be run even on a CPU. In contrast, MEHL-Soup+  still remains memory-efficient and can be run on a single GPU.

\subsection{Further Comparison between Greedy and Learned Soups}
\label{sec:further_comparison}

Apart from comparisons of efficiency and performance between different model soup methods, here we delve further into a comprehensive comparison between MEHL-Soup+ and Greedy-Soup from 
two perspectives: (i) total fine-tuning cost, and (ii) sensitivity to top-performing models,
which have not been explored in the previous literature yet are of importance for practical usage. 

\noindent
\textbf{Test accuracy vs total fine-tuning cost.}
In previous comparisons, we mainly focus on the model soup stage, where all fine-tuned models are already obtained. 
However, in many real-world scenarios, these models are fine-tuned with different hyperparameter configurations determined by grid/random search. Typically, the time cost of fine-tuning a single model is much larger than that of the model soup construction. For example, fine-tuning a CLIP ViT-B/32 on ImageNet requires around 4 GPU hours, while model soup training takes less than 1 hour. This is because fine-tuning is performed
on the training set, which is usually much larger than the validation set used in model soup training.
Thus, it is necessary to take the cost of fine-tuning stage into consideration. 

To investigate this, we gradually increase the number of fine-tuned models and measure the test accuracies achieved by Greedy-Soup and our MEHL-Soup+, as well as the corresponding total fine-tuning costs. 
Results are shown in 
\begin{wrapfigure}{r}{0.4\textwidth}
    \centering
    \includegraphics[width=1\linewidth]{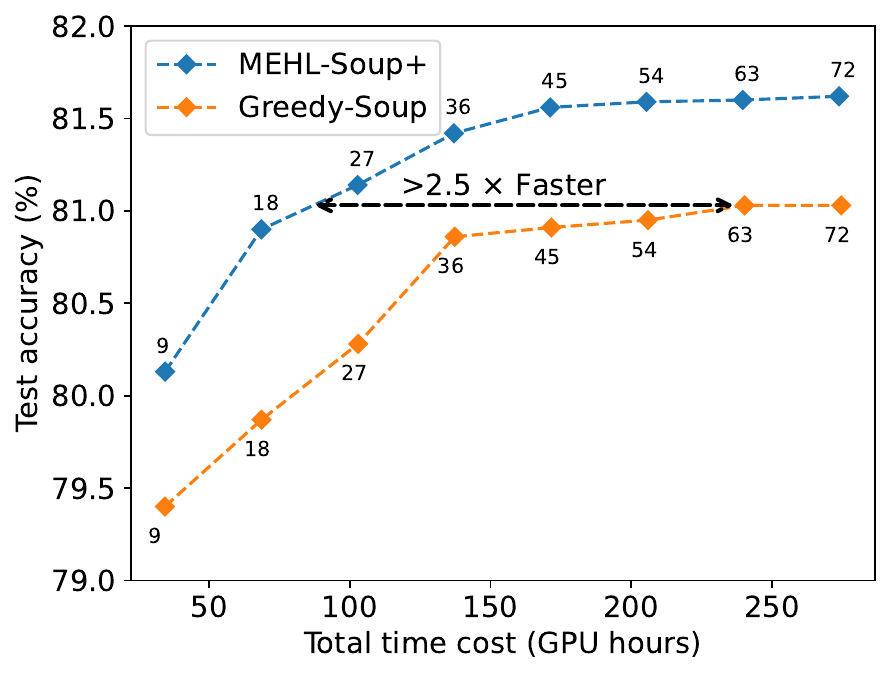}
    \caption{Test accuracy comparison of Greedy-Soup and MEHL-Soup+ w.r.t. fine-tuning time cost. 
    The experiment is performed on ImageNet with CLIP ViT-B/32. We use different numbers of fine-tuned models (displayed near the points) and measure their corresponding fine-tuning time costs. The model sequence follows the original random search order provided in \cite{wortsman2022model}. 
    }
    \label{fig:inc-num}
\end{wrapfigure}
\cref{fig:inc-num}.  It is evident that to reach comparable 
test accuracy, MEHL-Soup+ requires significantly fewer GPU hours than Greedy-Soup. 
For example, to attain a 
test accuracy
of 81\%, Greedy-Soup requires over 200 GPU hours while MEHL-Soup+ takes 
fewer than 100 GPU hours, a more than $2.5\times$ reduction. 
This is because MEHL-Soup+ achieves this performance with fewer than 27 fine-tuned models, thanks to its ability  to perform 
layer-wise 
weighted 
averaging with much better performance. In contrast, Greedy-Soup requires more than 54 models to achieve similar results.
Thus, such efficiency in the fine-tuning cost of MEHL-Soup+ brings further efficiency over Greedy-Soup beyond soup construction. 

\noindent
\textbf{Sensitivity to top-performing models.} 
To initialize
the greedy soup,
one selects the fine-tuned model with the best validation performance.
The remaining fine-tuned models are then sequentially tried to be added to the soup following a decreasing order of validation performance. 
In practice, grid/random search is commonly employed to identify the best fine-tuning hyperparameters~\cite{wortsman2022model}. However, obtaining a high-performance model through these search methods is often a challenging task that necessitates numerous trials. 
Thus, it is important
to examine whether such top-performing models are important to the success of greedy soup. 

\begin{table}[!t]
    \centering
	\setlength{\tabcolsep}{4pt}
    \caption{Test accuracies (\%) of different model soup methods after eliminating different numbers of top-performing models. We sort the fine-tuned CLIP ViT-B/32 models on ImageNet (\cref{sec:main-exp}) according to the validation accuracy and compare the performance of Greedy-Soup and MEHL-Soup+ after eliminating 2, 22, and 42 top-performing models.}
    % \vskip -.15in
    \footnotesize
    \label{tab:del_good_models}
    \begin{tabular}{@{}llllll@{}}
    \toprule
         models eliminated &- &Top-2 &Top-22 &Top-42  \\
    \midrule
         Greedy-Soup &81.03 &80.78 &80.08 &79.70\\
         MEHL-Soup+ &81.62{\scriptsize \color{teal} ($+0.59$)} &81.58{\scriptsize \color{teal} ($+0.80$)} &81.44{\scriptsize \color{teal} ($+1.36$)} &81.01{\scriptsize \color{teal} ($+1.31$)} \\
    \bottomrule
    \end{tabular}
    % \vskip -.15in
\end{table}

To this end, we replicate the ImageNet experiment in \cref{sec:main-exp} and sort the models in decreasing order of validation accuracy. We then compare the performance of Greedy-Soup and our MEHL-Soup+ after eliminating different numbers of top-performing fine-tuned models.  
From the results in \cref{tab:del_good_models},
we can observe that Greedy-Soup is more sensitive to
the top-performing models than MEHL-Soup+.
For example, after eliminating the top-2 performance models, Greedy-Soup experiences a significant drop 
of 0.25\% in accuracy,
while MEHL-Soup+ shows only a negligible drop. As elimination progresses to 22 and 42 top-performing models, the advantage of MEHL-Soup+ over Greedy-Soup becomes even more pronounced, resulting in an accuracy gain of 1.36\%. Remarkably, even after removing the top-42 performance models, MEHL-Soup+ still achieves an impressive 81\% test accuracy. 
These findings highlight that MEHL-Soup+ exhibits a significantly lower reliance on top-performing models thanks to the better flexibility by learned mixing coefficients.
This low reliance 
holds significant practical value and can help save some effort for hyperparameter search.

\subsection{Ablation Studies}
\label{sec:ablation}

\begin{figure}[!t]
	\centering
	\begin{subfigure}{0.25\linewidth}
		\centering
	\includegraphics[width=1\linewidth]{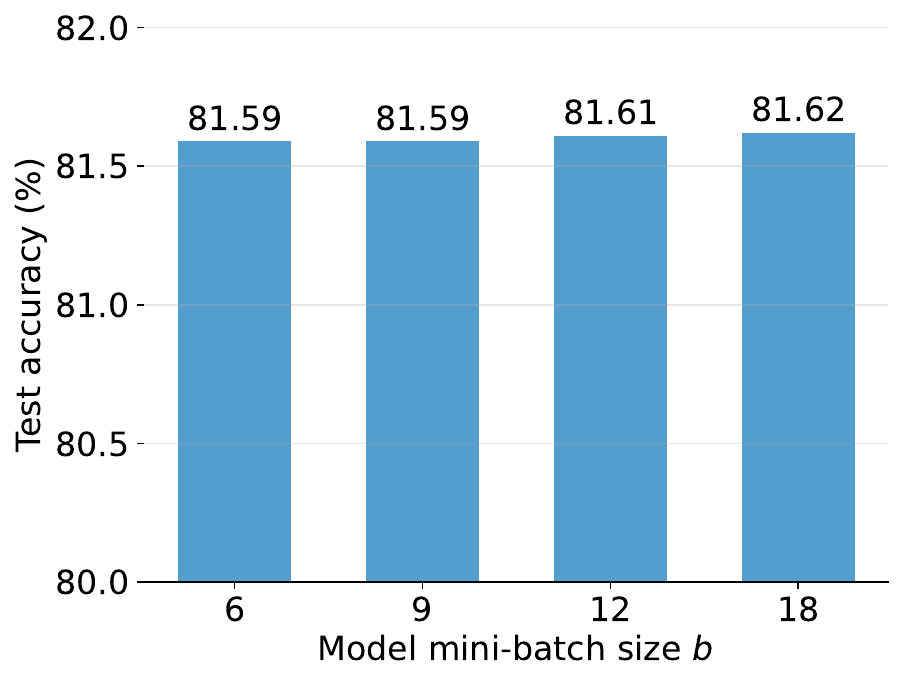}
  \caption{} \label{fig:ablation1}
	\end{subfigure}
\hfill
	\begin{subfigure}{0.25\linewidth}
		\centering
	\includegraphics[width=1\linewidth]{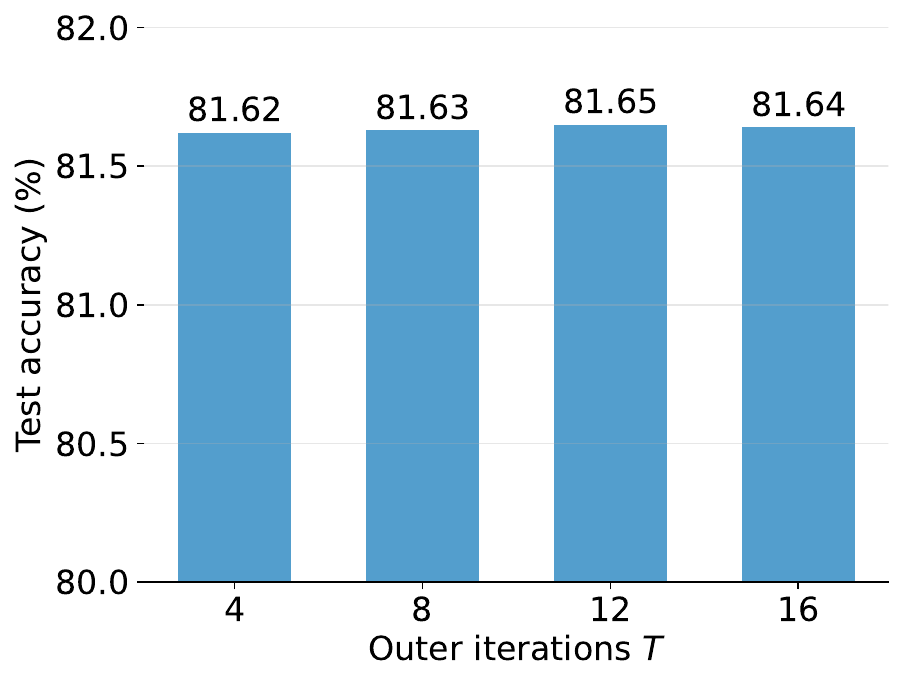}
  \caption{}\label{fig:ablation2}
	\end{subfigure}
\hfill
	\begin{subfigure}{0.25\linewidth}
		\centering
	\includegraphics[width=1\linewidth]{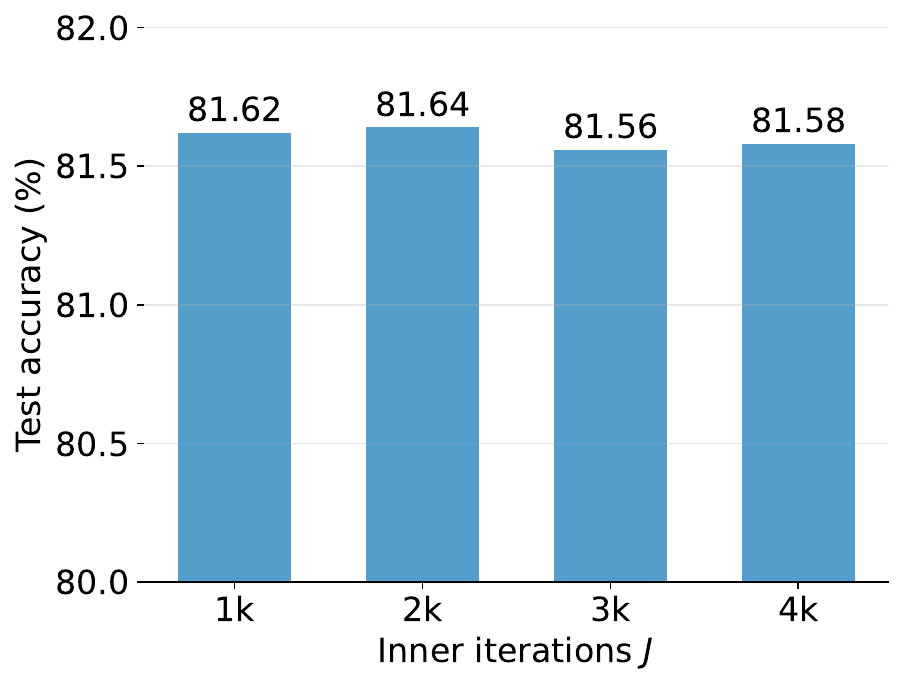}
  \caption{}\label{fig:ablation3}
	\end{subfigure}
	% \hspace{0.02in}
        % \vskip -.15in
	\caption{Sensitivity analysis of the hyperparameters in MEHL-Soup+. The experiments are conducted on ImageNet with CLIP ViT-B/32.} 
	\label{fig:abaltion}
 % \vskip -.2in
\end{figure}

\noindent
\textbf{Model mini-batch size.}
We first investigate the effects of model mini-batch size $b$ in MEHL-Soup+. \cref{fig:ablation1}
shows that the accuracy does not vary too much (<0.1\%) as the model mini-batch size varies.
In practice, one can adjust the model mini-batch size based on the available memory of the training device.

\noindent
\textbf{Number of model training iterations.}
Here we investigate how 
$T$,
the number of outer iterations,
affects accuracy. In \cref{fig:ablation2}, we observe that as $T$
increases, the performance gain becomes minor. Therefore, in our  experiments 
(\cref{sec:main-exp,sec:vit-L14,sec:further_comparison}),
we simply use one model training epoch (i.e. $\left \lceil K/b \right \rceil$) for efficiency.

\noindent
\textbf{Number of inner training iterations.} 
In \cref{sec:main-exp,sec:vit-L14,sec:further_comparison}, 
we 
use 1K inner iterations 
(corresponding to $5$ epochs over the validation set).
We further try 2K, 3K, and 4K iterations in this ablation study. As shown in \cref{fig:ablation3}, using more inner iterations does not yield a significant performance gain.
% and may even result in a slight decrease in test accuracy.

\section{Conclusion}
In this paper, we studied the scaling issue of learning mixing coefficients to build a model soup from numerous fine-tuned models.
We proposed a novel approach MEHL-Soup(+) based on
efficient hyperplane optimization and block coordinate gradient descent.  
MEHL-Soup(+) is computation- and memory-efficient and can be run on a single GPU.
Moreover, our method allows for extrapolated coefficients, and thus is more expressive than Learned-Soup whose coefficients are constrained in the probability simplex. 
We also theoretically established the convergence of MEHL-Soup(+).
Experimental results on various datasets demonstrate that MEHL-Soup(+) is more efficient and accurate than the Learned-Soup.
Furthermore, we hope our strategy could be beneficial to the fine-tuning process by unifying the fine-tuning and model averaging steps under a unified and systematic framework, which would lead to more efficient results.

% \newpage
\change{
\section*{Acknowledgements}
We thank anonymous reviewers for their valuable and constructive comments.
This work was partially supported by the National Key Research Development Project (2023YFF1104202), National Natural Science Foundation of China (62376155), Shanghai Municipal Science and Technology Research Program\,(2251 1105600)\, Major Project (2021SHZDZX0102), and the Research Grants
Council of the Hong Kong Special Administrative Region
(Grants 	C7004-22G-1 and 16202523).
}

% \newpage
\bibliographystyle{splncs04}
\bibliography{main}

\appendix
\renewcommand{\thetable}{A\arabic{table}}
\section{Proofs}
\label{apd:proof}
\begin{proof}[Proof of Theorem 3]
    At each outer iteration $t$, let $\hK_t=\{t_1,\dots, t_b\}$ be the indices of models being chosen.
    For $j=1,\dots, J$, 
    it follows from Taylor approximation that ($i=(t-1)J + j$),
    \begin{align}
        &  \hL(\valpha_{\cdot, i+1}; \hD^{\text{vl}})  \nonumber \\
        &= 
        \hL(\valpha_{\cdot, i}; \hD^{\text{vl}}) + \nabla_{\valpha_{\cdot, i}}^\top\hL(\valpha_{\cdot, i}; \hD^{\text{vl}})
        (\valpha_{\cdot, i+1}-\valpha_{\cdot, i}) \nonumber\\
        &\quad  + \frac{1}{2} (\valpha_{\cdot, i+1}-\valpha_{\cdot, i})^\top \nabla_{\valpha_{\cdot, i}}^2\hL(\vxi_i; \hD^{\text{vl}}) (\valpha_{\cdot, i+1}-\valpha_{\cdot, i}) \label{eq:temp-xas}\\
        & \leq  \hL(\valpha_{\cdot, i}; \hD^{\text{vl}}) 
        -\eta \nabla_{\valpha_{\cdot, i}}^\top\hL(\valpha_{\cdot, i}; \hD^{\text{vl}})
        \vU_i \nabla \hL(\valpha_{\cdot, i}; \hB_i) \nonumber \\
        & \quad  
        + \frac{\eta^2 \beta}{2} 
         \nabla^\top \hL(\valpha_{\cdot, i}; \hB_i) \vU_i^2 \nabla \hL(\valpha_{\cdot, i}; \hB_i)
        \label{eq:temp-asj}\\
        & =  \hL(\valpha_{\cdot, i}; \hD^{\text{vl}}) 
        -\eta \sum_{k=1}^{K} \bI(k\in \hK_t)\nabla_{\alpha_{k, i}}\hL(\alpha_{k, i}; \hD^{\text{vl}}) \nabla_{\alpha_{k, i}} \hL(\alpha_{k, i}; \hB_i) \nonumber \\
        &  \quad 
        + \frac{\eta^2 \beta}{2} 
        \sum_{k=1}^{K} \bI(k\in \hK_t) \left(\nabla_{\alpha_{k, i}} \hL(\alpha_{k, i}; \hB_i)\right)^2 \label{eq:temp-asm}
    \end{align}
    where $\vxi_i \in [\valpha_{\cdot, i}, \valpha_{\cdot, i+1}]$ in Eq.~\eqref{eq:temp-xas},
    $\vU_i = \text{diag}([\bI(1\in \hK_t), \bI(2\in \hK_t), \dots, \bI(K\in \hK_t) ]) \in \bR^{K\times K}$ (the $k$th diagonal entry indicates whether the $k$th model is chosen at iteration $i \in \{t(J-1), \dots, tJ\}$),
    Eq.~\eqref{eq:temp-asj} follows from the smoothness assumption and the update rule of $\valpha_{\cdot, i+1}$.
    Taking expectation w.r.t. $\hB_i$ on both sides of Eq.~\eqref{eq:temp-asm},
    we obtain
    \begin{align}
        &\bE_{\hB_i} \hL(\valpha_{\cdot, i+1}; \hD^{\text{vl}}) \nonumber\\
        &\leq \hL(\valpha_{\cdot, i}; \hD^{\text{vl}}) 
        -\eta \sum_{k=1}^{K} \bI(k\in \hK_t)\left(\nabla_{\alpha_{k, i}}\hL(\alpha_{k, i}; \hD^{\text{vl}}) \right)^2 \nonumber\\
        &\quad + \frac{\eta^2 \beta}{2} 
        \sum_{k=1}^{K} \bI(k\in \hK_t)\left( \left(\nabla_{\alpha_{k, i}} \hL(\alpha_{k, i}; \hD^{\text{vl}})\right)^2 + \textsf{Var}_{\hB_i}(\nabla_{\alpha_{k, i}} \hL(\alpha_{k, i}; \hB_i))\right) \label{eq:temp-axt} \\
        & \leq \hL(\valpha_{\cdot, i}; \hD^{\text{vl}}) 
        -\frac{\eta}{2} \sum_{k=1}^{K} \bI(k\in \hK_t)\left(\nabla_{\alpha_{k, i}}\hL(\alpha_{k, i}; \hD^{\text{vl}}) \right)^2 \nonumber\\
        &\quad + \frac{\eta^2 \beta}{2} 
        \sum_{k=1}^{K} \bI(k\in \hK_t) \textsf{Var}_{\hB_i}(\nabla_{\alpha_{k, i}} \hL(\alpha_{k, i}; \hB_i))  \label{eq:temp-aps}
    \end{align}
    where $\textsf{Var}(x)$ is the variance of $x$,  Eq.~\eqref{eq:temp-axt} follows from the identity
    $\bE x^2 =( \bE x)^2 + \textsf{Var}(x)$,
    and
    Eq.~\eqref{eq:temp-aps} 
    follows from the assumption $\eta \leq \frac{1}{\beta}$ (thus, $\frac{\eta^2\beta}{2} \leq \frac{\eta}{2}$).
    Taking expectation w.r.t. $\valpha_{\cdot, i}$ and $\valpha_{\cdot, i+1}$ on both sides of Eq.~\eqref{eq:temp-aps},
    we have
    \begin{align}
        \! \bE_{\valpha_{\cdot, i+1}} \hL(\valpha_{\cdot, i+1}; \hD^{\text{vl}}) \!
        & \leq \bE_{\valpha_{\cdot, i}}  \! \hL(\valpha_{\cdot, i}; \hD^{\text{vl}}) 
        \!-\!\frac{\eta}{2} \! \sum_{k=1}^{K} \! \bI(k\in \hK_t) \bE_{\alpha_{k, i}}  \!\!\left(\nabla_{\alpha_{k, i}}\hL(\alpha_{k, i}; \hD^{\text{vl}}) \right)^2 \nonumber\\
        & \quad + \frac{\eta^2 \beta}{2} 
        \sum_{k=1}^{K} \bI(k\in \hK_t) \bE_{\alpha_{k, i}}  \textsf{Var}_{\hB_i}(\nabla_{\alpha_{k, i}} \hL(\alpha_{k, i}; \hB_i)) 
    \end{align}
    Summing over $j=1 \to J$ and $t=1\to T$, let $i=(t-1)J + j$, we obtain
    \begin{align}
        &\bE_{\valpha_{\cdot, TJ}} \hL(\valpha_{\cdot, TJ}; \hD^{\text{vl}}) \nonumber \\
        & \leq  \bE_{\valpha_{\cdot, 1}}  \hL(\valpha_{\cdot, 1}; \hD^{\text{vl}}) 
        -\frac{\eta}{2} \sum_{t=1}^T \sum_{j=1}^{J} \sum_{k=1}^{K} \bI(k\in \hK_t) \bE_{\alpha_{k, i}}  \left(\nabla_{\alpha_{k, i}}\hL(\alpha_{k, i}; \hD^{\text{vl}}) \right)^2 \nonumber \\ 
        &\quad + \frac{\eta^2 \beta}{2} 
         \sum_{t=1}^T \sum_{j=1}^{J}\sum_{k=1}^{K} \bI(k\in \hK_t) \bE_{\alpha_{k, i}}  \textsf{Var}_{\hB_i}(\nabla_{\alpha_{k, i}} \hL(\alpha_{k, i}; \hB_i)) 	\nonumber	\\ %newline
            &=  \bE_{\valpha_{\cdot, 1}}  \hL(\valpha_{\cdot, 1}; \hD^{\text{vl}}) 
         -\frac{\eta}{2} \sum_{t=1}^T \sum_{k=1}^{K} \bI(k\in \hK_t) \left(\sum_{j=1}^{J}  \bE_{\alpha_{k, i}}  \left(\nabla_{\alpha_{k, i}}\hL(\alpha_{k, i}; \hD^{\text{vl}}) \right)^2\right) \nonumber\\
         &\quad  + \frac{\eta^2 \beta \sigma^2 TJ}{2} \label{eq:temp-auq}	\\ %newline
         &\leq  \bE_{\valpha_{k, 1}}  \hL(\valpha_{\cdot, 1}; \hD^{\text{vl}}) 
         -\frac{\eta}{2} \sum_{t=1}^T \sum_{k=1}^{K} \bI(k\in \hK_t)  \bE_{\alpha_{k, tJ}}  \left(\nabla_{\alpha_{k, tJ}}\hL(\alpha_{k, tJ}; \hD^{\text{vl}}) \right)^2  \nonumber\\
         &\quad  + \frac{\eta^2 \beta \sigma^2 TJ}{2}  \label{eq:temp-asq}
    \end{align}
    where
    Eq.~\eqref{eq:temp-auq} follows from $\sum_{k=1}^{K} \bI(k\in \hK_t) \bE_{\alpha_{k, i}}  \textsf{Var}_{\hB_i}(\nabla_{\alpha_{k, i}} \hL(\alpha_{k, i}; \hB_i)) \leq \sigma^2$,
    Eq.~\eqref{eq:temp-asq}
    follows from $$\sum_{j=1}^{J}  \bE_{\alpha_{k, i}}  \left(\nabla_{\alpha_{k, i}}\hL(\alpha_{k, i}; \hD^{\text{vl}}) \right)^2 \geq  \bE_{\alpha_{k, tJ}}  \left(\nabla_{\alpha_{k, tJ}}\hL(\alpha_{k, tJ}; \hD^{\text{vl}}) \right)^2.$$
    Taking expectation w.r.t. $\bI(k\in \hK_t)$, we have
    \begin{align}
        &\bE_{\valpha_{\cdot, TJ}} \hL(\valpha_{\cdot, TJ}; \hD^{\text{vl}}) \nonumber\\
        & \leq     \bE_{\valpha_{\cdot, 1}}  \hL(\valpha_{\cdot, 1}; \hD^{\text{vl}}) 
        - \frac{\eta b}{2K}  \sum_{t=1}^T \bE_{\valpha_{\cdot, tJ}}  \|\nabla_{\valpha_{\cdot, tJ}}\hL(\valpha_{\cdot, tJ}; \hD^{\text{vl}}) \|^2   + \frac{\eta^2 \beta \sigma^2 TJ}{2} \label{eq:temp-asy}	 \\
        & \leq     \bE_{\valpha_{\cdot, 1}}  \hL(\valpha_{\cdot, 1}; \hD^{\text{vl}}) 
        \!-\! \frac{\eta Tb}{2K} \min_{1\leq t \leq T}\bE_{\valpha_{\cdot, tJ}}  \|\nabla_{\valpha_{\cdot, tJ}}\hL(\valpha_{\cdot, tJ}; \hD^{\text{vl}}) \|^2  \!+\! \frac{\eta^2 \beta \sigma^2 TJ}{2}\!\!
        \label{eq:temp-asdy}
    \end{align}
    where Eq.~\eqref{eq:temp-asy} follows from $\bE \bI(k\in \hK_t) = \frac{b}{K}$
    and Eq.~\eqref{eq:temp-asdy} follows from 
    $$
    \min_{1\leq t \leq T}\bE_{\valpha_{\cdot, tJ}}  \|\nabla_{\valpha_{\cdot, tJ}}\hL(\valpha_{\cdot, tJ}; \hD^{\text{vl}}) \|^2 \leq \sum_{t=1}^T  \bE_{\valpha_{\cdot, tJ}}  \|\nabla_{\valpha_{\cdot, tJ}}\hL(\valpha_{\cdot, tJ}; \hD^{\text{vl}}) \|^2.
    $$
    Rearranging Eq.~\eqref{eq:temp-asdy}, it follows that
    \begin{align}
        & \frac{\eta Tb}{2K} \min_{1\leq t \leq T} \bE_{\valpha_{\cdot, tJ}}  \|\nabla_{\valpha_{\cdot, tJ}}\hL(\valpha_{\cdot, tJ}; \hD^{\text{vl}}) \|^2   \nonumber\\
        &\leq \bE_{\valpha_{\cdot, 1}}\hL(\valpha_{\cdot, 1}; \hD^{\text{vl}}) - \bE_{\valpha_{\cdot, TJ}}  \hL(\valpha_{\cdot, TJ}; \hD^{\text{vl}})  + \frac{\eta^2 \beta \sigma^2 TJ}{2}  \label{eq:temp-asi}
    \end{align}
    Let $\eta=\text{min}\{\frac{1}{\sqrt{T}}, \frac{1}{\beta}\}$, from Eq.~\eqref{eq:temp-asi}, we have,
    \begin{align}
        &\frac{b \sqrt{T}}{2K} \min_{1\leq t \leq T} \bE_{\valpha_{\cdot, tJ}}  \|\nabla_{\valpha_{\cdot, tJ}}\hL(\valpha_{\cdot, tJ}; \hD^{\text{vl}}) \|^2   \nonumber\\
        &\leq \bE_{\valpha_{\cdot, 1}} \hL(\valpha_{\cdot, 1}; \hD^{\text{vl}}) - \bE_{\valpha_{\cdot, TJ}}  \hL(\valpha_{\cdot, TJ}; \hD^{\text{vl}}) +\frac{ \beta  J \sigma^2}{2}
    \end{align}
    Diving both sides of the above inequality by $\frac{b \sqrt{T}}{2K}$, we obtain
    \begin{align}
        & \min_{1\leq t \leq T} \bE_{\valpha_{\cdot, tJ}}  \|\nabla_{\valpha_{\cdot, tJ}}\hL(\valpha_{\cdot, tJ}; \hD^{\text{vl}}) \|^2   \nonumber\\
        &\leq \frac{2K\left(\bE_{\valpha_{\cdot, 1}} \hL(\valpha_{\cdot, 1}; \hD^{\text{vl}}) - \bE_{\valpha_{\cdot, TJ}}  \hL(\valpha_{\cdot, TJ}; \hD^{\text{vl}})\right)}{b\sqrt{T}} +  \frac{\beta J  \sigma^2 K}{b\sqrt{T}},
    \end{align}
    and we finish the proof.
\end{proof}

\section{Training Details}

\emph{Model Fine-tuning.}
The CLIP ViT-B/32 models fine-tuned on ImageNet are publicly available and can be found in \url{https://github.com/mlfoundations/model-soups}. There are 72 models fine-tuned with a random hyperparameter search over the learning rate, weight decay, training epochs, label smoothing, and data augmentation. The CLIP ViT-L/14 models are fine-tuned with random search over learning rates of $\{1\times10^{-6}, 3\times10^{-6}, 5\times10^{-6}, 1\times10^{-5}, 2\times10^{-5}, 3\times10^{-5}\}$ and weight decays of $\{0,  0.001, 0.003, 0.005, 0.01, 0.03\}$. We fine-tune 32 models for CIFAR-10/100 and 8 models for ImageNet, as the latter requires substantially higher computational costs for fine-tuning. We use a batch size of 128 and set the training epoch to 10. All training images are resized to $224\times224$ with standard CLIP-style image preprocessing~\cite{radford2021learning}, while for ImageNet, we additionally apply RandAugmentation~\cite{cubuk2020randaugment} following~\cite{wortsman2022model}. We use a random split of 10\% training data as the held-out validation set for CIFAR-10/100 and 2\% for ImageNet. We will release the fine-tuned model checkpoints.

\noindent
\emph{Learned Soup Training.}
We train HL-Soup(+) and MEHL-Soup(+) for 1 model training epochs, i.e., loading  $K$ models into memory for once, and use a model batch size of $b=18$ for CLIP ViT-B/32 and $b=8$ for CLIP ViT-L/14. This corresponds to outer iterations of  $T=\lceil K/b \rceil$.
We use a batch size of $128$ for training. 
To accommodate ViT-L/4 and its memory requirements within a 24GB GPU, we employ 8 times of gradient accumulation.
% During training,
% we use a batch size of $128$, and adopt 8 times gradients accumulation
% \footnote{*** write more clearly what u mean by "8 times gradients accumulation"} 
% for ViT-L/14 to fit into 24GB GPU memory. 
For HL-Soup(+), MEHL-Soup(+), and Learn-Soup(+), we use 1K inner iterations and adopt the AdamW optimizer with a cosine learning rate schedule.
The learning rate is searched over $\{0.005, 0.01, 0.05\}$ and we choose $0.01$ for  HL-Soup(+) and MEHL-Soup(+), and $0.05$ for  Learn-Soup(+). Weight decay is searched over  $\{0, 0.001, 0.01, 0.05, 0.1, 0.5\}$ and we use $0.1$ for  HL-Soup(+) and MEHL-Soup(+), and $0$ for  Learn-Soup(+) for optimal.

\section{More Results}
To further evaluate the performance of our approach on different architectures, we conduct an experiment on CIFAR-100 with 32 models fine-tuned from ResNet-101 (pre-trained on ImageNet). The setting is the same as in Tab.~2.
The table below shows that MEHL-Soup+ is more efficient and effective than Learned-Soup+. 
\begin{table}[htbp]
% \vskip -.15in
    \centering
    % \tiny
    % \small
    % \setlength{\tabcolsep}{1mm} 
    % \resizebox{.44\textwidth}{!}{
    \caption{Comparison of different methods with pre-trained ResNet-101 on CIFAR-100. The number of fine-tuned models is 32. We measure the time and memory on a server with one NVIDIA GeForce RTX 4090 GPU and 256 GB RAM.}
    \begin{tabular}{lccccc}
    \toprule
    % \hline
     % Method & Test Acc &Time &Ep. &Cons. Time &Memory \\
         &Testing  &Time & &Soup &Peak \\
     Method &accuracy &per &\#epoch &construction &memory   \\
     &(\%) &epoch & &time &burden \\
    \midrule
    % \hline
    Best individual     &87.10 &- &- &- &-\\
    \midrule
    Uniform-Soup~\cite{izmailov2018averaging} &86.47 &- &- &- &-\\
    Greedy-Soup~\cite{wortsman2022model}  &88.62 &5s &63 &328s &4GB\\
        \arrayrulecolor{lightgray}
    \midrule
    Learned-Soup~\cite{wortsman2022model}  &88.50 &913s &5 &4566s &68GB\\
    HL-Soup~(\textbf{ours}) &88.82 &47s &5 &238s &17GB\\
    MEHL-Soup~(\textbf{ours}) &89.03 &10s &20 &207s &13GB\\
    \midrule
    Learned-Soup+~\cite{wortsman2022model} &88.83 &915s &5 &4576s &68GB\\
    HL-Soup+~(\textbf{ours}) &89.12 &48s &5 &244s &17GB\\
    MEHL-Soup+~(\textbf{ours}) &\textbf{89.32} &10s &20 &209s~{ \color{teal} \footnotesize ($\downarrow 21.9\times$)} &13GB~{\color{teal} \footnotesize ($\downarrow 5.2\times$)}\\
    % \hline
        \midrule
    Ensemble &89.61 &- &- &- &-\\
    %      & \\
        \arrayrulecolor{black}
    \bottomrule
    \end{tabular}
    % }
    % \caption{Caption}
    \label{tab:my_label}
    % \vspace{-4mm}
\end{table}

\section{Ablation on Weight Decentralization}
Learning with $\boldsymbol{\theta}_k\!-\!\bar{\boldsymbol{\theta}}$, i.e., weight decentralization, as in Eq.~(3) can be more effective and produce better performance.
We conduct an ablation (with the same setting as in Tab.~1) to compare MEHL-Soup+ with simply using $\vtheta_{\star}'\!=\sum_{k=1}^K\alpha_k\vtheta_k$ (without weight decentralization). 
The test accuracy drops significantly from 81.62 to 81.02. This is perhaps because $\{\boldsymbol{\theta}_k\}_{k=1}^K$ are in the same basin and decomposing them as 
$\{\vtheta_k - \bar{\vtheta}\}_{k=1}^K$ can reduce correlation (the averaged cosine similarity between different weight vectors is decreased from 0.99 to 0.19) for better optimization.

\end{document}